# CREATING LIGHTWEIGHT OBJECT DETECTORS WITH MODEL COMPRESSION FOR DEPLOYMENT ON EDGE DEVICES

*Yiwu Yao, Weiqiang Yang, Haoqi Zhu*

NetEase Ltd. Hangzhou Research Institute, Hangzhou, 310052, China
{yaoyiwu, yangweiqiang, zhuhaoqi}@corp.netease.com

## ABSTRACT

To achieve lightweight object detectors for deployment on the edge devices, an effective model compression pipeline is proposed in this paper. The compression pipeline consists of automatic channel pruning for the backbone, fixed channel deletion for the branch layers and knowledge distillation for the guidance learning. As results, the Resnet50-v1d is auto-pruned and fine-tuned on ImageNet to attain a compact base model as the backbone of object detector. Then, lightweight object detectors are implemented with proposed compression pipeline. For instance, the SSD-300 with model size=16.3MB, FLOPS=2.31G, and mAP=71.2 is created, revealing a better result than SSD-300-MobileNet.

## 1. INTRODUCTION

CNN-based object detectors have been widely applied in various fields, such as video surveillance, advanced driving assistant systems (ADAS) and medical image analysis, etc. The framework of object detection is mainly divided into two categories, including one-stage method [1-4] and two-stage method [5] [6]. The two-stage detectors with region proposal network (RPN) can effectively facilitate the class imbalance, thus achieve remarkable performance on major benchmarks, such as Pascal VOC and MS COCO. However, to deploy the two-stage detectors on edge devices is unrealistic due to the disadvantage of inference speed. To accelerate detection, one-stage detectors are investigated, where the RPN is discarded to construct a fully convolutional single shot dataflow.

To further improve the forward efficiency of one-stage detectors, especially for deployment on edge devices, model compression techniques are usually introduced to reduce the model capacity and computation complexity. Approaches for model compression can be classified into the following areas: network pruning [7-11], model quantization [12-14], network simplification [15-17] and knowledge distillation [18-20], etc. Network pruning aims to reduce parameter redundancy by inducing model sparsity, while model quantization achieves regularization by clustering parameters and activations onto discrete and reduced-precision points. The simplification of network structure targets at obtaining efficient CNN models, such as MobileNet, ShuffleNet and MnasNet, etc. Knowledge distillation assisted with the soft labels and guidance features from teacher network, can effectively supervise the learning of student network. In order to generate lightweight object detectors, the model compression can be utilized in two ways: on one hand, object detectors can be designed with simplified networks as backbone, such as SSD-MobileNet; on the other hand, object detectors with large model size can be pruned or quantized into efficient ones, taking account into the tradeoff between accuracy and efficiency.

In this paper, by jointly exploiting model compression techniques, a lightweight object detector is created based on the framework of single shot multi-box detector (SSD). The concrete procedure consists of following aspects: First, the Resnet50-v1d is pruned and fine-tuned on ImageNet dataset to attain a simplified network as the backbone of detection model. Second, simple residual blocks are added to detection branches. Third, branch layers except backbone are pruned with a fixed prune-rate, then randomly initialized for further training. In addition, knowledge distillation is adopted for better supervision of training on detection task. Consequently, a lightweight SSD-300 is obtained with model size=16.3MB, FLOPS=2.31G, and mAP=71.2, indicating a better result than SSD-300-MobileNet.

This paper is organized as follows: Section 2 elaborates the compression pipeline for creating lightweight detection model, including automatic channel pruning, fixed channel deletion for branch layers and knowledge distillation. Section 3 gives the experimental results on created object detectors, followed by conclusions in section 4.

## 2. MODEL COMPRESSION PIPELINE FOR DETECTION MODELS

In this section, the model compression pipeline specific for realizing lightweight detection models is clarified in detail.

### 2.1. Overall Framework of the Compression Pipeline

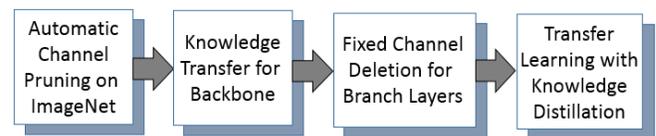

**Fig. 1.** Overall framework of the model compression pipeline

The overall framework of proposed compression pipeline for creating lightweight object detectors is shown in Fig. 1. First, the Resnet50-v1d is auto-pruned and fine-tuned on ImageNet to explore a simplified base network. Then, the pruned CNN serves as the backbone of designed object detector. After that, heavy branch layers are pruned with a fixed prune-rate, and then randomly initialized. At last, the knowledge distillation technique can be leveraged to further improve the accuracy of compressed detection model.

Resnet50-v1d with post-activation bottleneck structure is preferred as the backbone of SSD, since 7×7 convolutional layer before the first residual block is substituted by three stacked 3×3 convolutional layers, leading to efficient and compact network architecture. Hence, the realized detector only contains 1×1 and 3×3 convolution operations, which can be readily supported by any off the shelf platforms, especially the ones with Winograd implementation.

## 2.2. Automatic Channel Pruning for Backbone

### 2.2.1. Pruning Strategy Based on Network Slimming

Channel pruning methods prune the filter weights of CNN at filter level, thus the compressed structure is compatible with existing deep learning frameworks or mature platforms. The pruning strategy can be roughly classified into the following three categories: The first method is that, given a pretrained model, the output channels in each layer are pruned according to filter importance, such as taylor expansion criteria [8]; The second method aims to minimize the reconstruction error of output features between pruned model and pretrained model, by layer-wisely selecting the most important channel subset for each layer, such as ThiNet [9] and channel selection [10]. The third pruning method firstly regularizes the model with sparse constraints at filter-level, and then prunes the sparse channels away with a specific pruning mask, such as network slimming [11]. Compared with the first and second pruning methods, in spite of extra training cost for sparse constraint, the regularized model has a lower risk of over-fitting, and the global pruning mask is easier to determine for achieving the pruning goal.

Attributed to Batch Normalization (BN) operation after each convolutional layer in Resnet50-v1d, network slimming can be very suitable for regularization and automatic channel pruning to yield compact CNN model. The concrete pruning strategy proposed in [11] contains three aspects, covering the regularization, automatic pruning and fine-tuning. During training, network slimming imposes simple $L_1$ regularization on channel-wise scaling factors within BN layers as below:

$$L = \sum_{(x,y)} l(f(x,W),y) + \lambda \sum_{\gamma \in \Gamma} g(\gamma) \qquad (1)$$

where $(x, y)$ denote the training sample and label, $W$ denotes the trainable weights, the term $l(\cdot)$ corresponds to the normal loss of a CNN, the term $g(\gamma) = |\gamma|$ is introduced as auxiliary loss to impose sparsity constraint on the gamma coefficients within BN layers, and $\lambda$ balances the two losses.

During pruning stage, the actual prune-rate for each layer is determined in line with the target prune-rate and the global ranking of $|\gamma|$. Attributed to such automatic pruning approach with varied prune-rate for each layer, the resulted compact model will have a more reasonable and effective structure, leading to moderate accuracy loss.

As recommended in [11] for Resnet trained on ImageNet dataset, the learning rate is initialized as 0.1, and is multiplied by 0.1 after every 30 epochs, and the balance factor $\lambda$ is set as $10^{-5}$. The network slimming with this original setting of hyper-params in [11] is denoted as Slim-A. In this paper, in order to achieve a more robust regularization effectiveness for Resnet50-v1d on ImageNet, the learning rate is decayed by using cosine annealing as below:

$$l_r = 0.5 l_{r,0} \left( cos \left( \pi \frac{t}{epoch_{max}} \right) + 1 \right) \qquad (2)$$

where $l_{r,0}$=0.1, $epoch_{max}$=120, $t$ denotes the current training epoch. In addition, the balance factor $\lambda$ is set as $10^{-4}$ to impose more strict penalty on scaling factors. The network slimming with new setting of hyper-params in this paper is denoted as Slim-B. Slim-A and Slim-B are comprehensively compared in experiments, and it is proven that the effectiveness of Slim-B outperforms Slim-A. Thus, Slim-B is preferred in model compression pipeline to yield the backbone of lightweight object detectors.

### 2.2.2. Automatic Pruning for Resnet50-v1d

In general, in order to guarantee the output channels to be matched among different residual blocks in the same group, the pruning mask for the bottleneck structure of Resnet keeps output channels of residual blocks unpruned. Nevertheless, pruning these output channels can further help to increase the compression ratio and speedup ratio, leading to more efficient compact base-models for transfer learning.

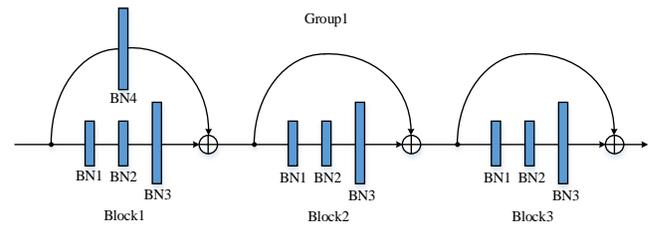

**Fig. 2.** The first residual block in Resnet50-v1d.

For instance, the first residual group of Resnet50-v1d is illustrated in Fig. 2, where only BN layers are depicted as the channel pruning masks. We denote $m_{i,j}$ as the channel mask of the $j$-th BN layer in the $i$-th residual block, then the output channels of involved three blocks can be matched on the basis of unified channel mask as below:

$$m = m_{1,3} | m_{1,4} | m_{2,3} | m_{3,3} \qquad (3)$$

where the symbol | represents logic *or* operation. By applying the unified mask to prune the output channels of each block,

unimportant output channels shared by these blocks can be effectively removed, while channel matching among these blocks is still guaranteed.

Afterwards, the channel matching strategy is extended to other residual groups. Meanwhile, output channels of other convolutional layers are normally pruned with channel masks determined by relevant followed BN layers. Consequently, owing to the global pruning and proposed channel matching strategy, automatic channel pruning can be conducted to yield a reasonable and effective pruned network structure.

### 2.3. Fixed Channel Deletion for Branch Layers

*2.3.1. Simple Residual Block Added to Detection Branch*

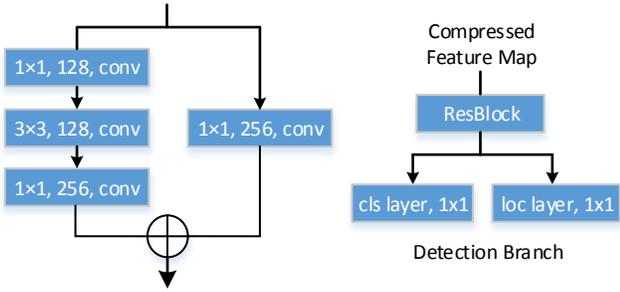

(a) Simple residual block     (b) Modified detection branch
**Fig. 3.** Simple residual block (denoted as ResBlock) with default output channel width, and the modified detection branch.

Based on the standard structure of SSD detector, the simple residual blocks with post-activation bottleneck structure (as shown in Fig. 3) are added before *cls* layers and *loc* layers within the detection branches, including the branches with input feature map sizes of 19×19, 10×10, 5×5, 3×3 and 1×1. In addition, the 1×1 convolutional layers are used to replace the original 3×3 *cls* and *loc* layers in SSD to reduce the model complexity.

Since the backbone is transferred from the pruned base model, and the extras layers will be further pruned to increase compression ratio (described in next subsection), the feature maps fed to detection branches are compressed along channel dimension. Thus, the added simple residual blocks can help to enhance the representation capability of compressed input feature maps, then improve the detection accuracy.

*2.3.2. Channel Pruning with Fixed Prune-rate*

The layers except the backbone in SSD are referred as branch layers, including the extras layers, ResBlock layers, *cls* and *loc* layers. Pruning the full or partial branch layers of the detector can obtain a remarkable reduction in model size and computation. Since the branch layers are randomly initialized, channels of these layers can be pruned with a proper fixed prune-rate before training.

In this paper, the output channels of ResBlock, *cls* layers and *loc* layers remain unchanged, and the input channels of the layers connected to backbone are determined by relevant pruned layers, while both input and output channels of other selected branch layers are pruned with a proper fixed prune-rate. Eventually, an efficient and effective lightweight SSD can be achieved with auto-pruned backbone and fixed-pruned branch layers.

### 2.4. Knowledge Distillation for Object Detector

The knowledge distillation (KD) for Faster-RCNN is firstly presented in [20], of which the soft-loss contains soft KD loss, teacher bounded $L_2$ regression loss and hint based $L_2$ loss. In this paper, the soft loss in [20] is refined to be adapt to guiding the transfer learning of created lightweight SSD.

First of all, the total loss with knowledge distillation for SSD is formulated as below:

$$L = \frac{1}{N}\sum_i L_{cls} + \alpha \frac{1}{N}\sum_j L_{loc} + \beta L_{AT} \quad (4)$$

where $L_{cls}$ is the *cls* loss, $L_{loc}$ is the *loc* loss, $L_{AT}$ is the attention transfer loss, $N$ is the number of recalled positives, and $\alpha, \beta$ are balance factors and both fixed as 1.0.

The $L_{cls}$ with hard loss and soft loss is denoted as below:

$$\begin{cases} L_{cls} = (1-\mu)L_{hard}(P_s, y_{cls}) + \mu L_{soft}(P_s, P_t) \\ L_{soft} = -\sum \omega_c P_t log(P_s) \end{cases} \quad (5)$$

where $L_{hard}$ is the hard loss using the true label $y_{cls}$ and *cls* predict of student ($P_s$), $L_{soft}$ is the soft KD loss [18] using the soft label of teacher ($P_t$) and $P_s$. Meanwhile, $\omega_c$ is set as 1.5 for background and set as 1.0 for other objects to facilitate the class imbalance, and the factor $\mu$ is initialized as 0.9 then multiplied by 0.5 after every 60 epochs during training.

The *loc* regression loss $L_{loc}$ is expressed as below:

$$L_{loc} = L_{smooth\_L1}(R_s, y_{loc}) + \nu L_b(R_s, R_t, y_{loc}) \quad (6)$$

where $L_{smooth\_L1}$ is the smooth $L_1$ loss using the true label $y_{loc}$ and *loc* predict of student ($R_s$), $L_b$ is the teacher bounded loss only applied on recalled positives, and the factor $\nu$ is set as 0.5. $L_b$ is calculated as $\|R_s - y_{loc}\|_2^2$, if $\|R_s - y_{loc}\|_2^2 + m > \|R_t - y_{loc}\|_2^2$, otherwise $L_b$ is zero, here $m$=1.5.

The last term $L_{AT}$ is adopted to replace the hint loss in [20], thus the adaption layers for hint learning are never used, resulting in reduction of training cost. More importantly, the guidance based on attention transfer can directly compare the difference between the compressed feature maps and original feature maps. The output feature maps of randomly initialized extras layers and simple ResBlocks are collected to compute attention maps, then the $L_{AT}$ can be derived as:

$$L_{AT} = \sum_i \left\| \frac{F(A_i^t)}{\|F(A_i^t)\|_2} - \frac{F(A_i^s)}{\|F(A_i^s)\|_2} \right\|_2 \quad (7)$$

where $A_i^t, A_i^s$ are the *i*-th feature map collected from teacher and student respectively, and $F(A) = \sum_{j=1}^{C}|A_j|^2$ is used [19] to compute attention maps.

## 3. EXPERIMENTAL RESULTS

### 3.1. Effectiveness of Automatic Channel Pruning

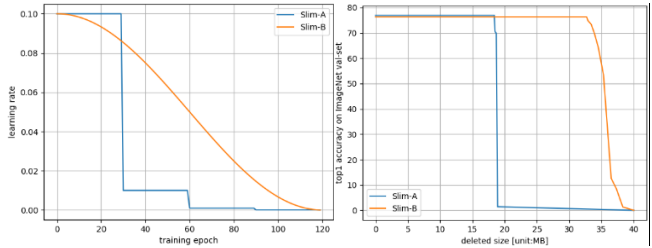

(a) Learning rate adjustment    (b) Accuracy vs. removed size
**Fig. 4.** Learning rate adjustment and regularization effectiveness.

**Table 1.** Regularization results of Resnet50-v1d using Slim-A and Slim-B on ImageNet-2012 validation-set: Original Model means the pre-trained Resnet50-v1d; Model-Slim-A means the model regularized with Slim-A; and Model-Slim-B means the model regularized with Slim-B.

| Model | Capacity | Top-1 Acc. | Top-5 Acc. |
|---|---|---|---|
| Original Model | 97.6MB | 76.9 | 93.6 |
| Model-Slim-A | 97.6MB | 76.5 | 93.2 |
| Model-Slim-B | 97.6MB | 76.3 | 93.1 |

**Table 2.** Fine-tuning results of three auto-pruned models using Slim-B on ImageNet-2012 validation-set: FLOPS is calculated as the sum of multiplies and adds in all layers; '-dt' means that the channel matching proposed in sec. 2.2.2 is not used.

| Model | Capacity | FLOPS | Top-1 | Top-5 |
|---|---|---|---|---|
| Resnet-A | 64.9MB | 4.71G | 76.2 | 92.9 |
| Resnet-B | 38.2MB | 2.89G | 74.7 | 92.1 |
| Resnet-C | 12.1MB | 1.17G | 68.8 | 88.9 |
| Resnet-C-dt | 24.3MB | 2.25G | 71.1 | 90.1 |

The decay curves of learning rate adjustment for Slim-A and Slim-B are depicted in Fig. 4 (a). It is obvious that the curve of Slim-A is steeper than Slim-B, where the learning rate decays faster for Slim-A after training 30 epochs. Hence, the regularization result of Resnet50-v1d using Slim-B is better than the one with Slim-A. From Fig. 4 (b) and Table. 1, it is seen that the accuracy of regularized model with Slim-B is very close to the one of regularized model with Slim-A, but the curve of accuracy vs. removed parameters with Slim-B is more robust than the one with Slim-A.

As results, three base models by using automatic channel pruning are achieved as listed in Table. 2, including Resnet-A, Resnet-B and Resnet-C. The accuracy losses of these three compact models are effectively recovered by fine-tuning 60 epochs, with the learning rate initialized as $10^{-2}$ and decayed to $10^{-3}$ after 30 epochs. Meanwhile, compared with Resnet-C-dt, the Resnet-C by channel matching possesses much less parameters while indicates sufficient precision. Eventually, the Resnet-C is selected for transfer learning to accomplish building the lightweight SSD (backbone size=11.6MB).

### 3.2. Results on Pascal VOC

**Table 3.** Results of created detectors on Pascal VOC07 test-set: Resnet-C is transferred to the backbone of object detectors; +P means all branch layers are pruned with fixed prune-rate; +R means simple ResBlocks are added to detection branches; +KD means knowledge distillation is used for guidance learning.

| Models | Input Size | Fixed-ratio | Capacity | FLOPS | mAP |
|---|---|---|---|---|---|
| +P | 300 | 0.4 | 19.6MB | 2.36G | 68.1 |
| +P+R | 300 | 0.4 | 18.9MB | 2.36G | 70.9 |
| +P+R+KD | 300 | 0.4 | 18.9MB | 2.36G | 71.7 |
| +P | 300 | 0.6 | 17.1MB | 2.34G | 67.1 |
| +P+R | 300 | 0.6 | 16.3MB | 2.31G | 69.9 |
| +P+R+KD | 300 | 0.6 | 16.3MB | 2.31G | 71.2 |
| +P | 512 | 0.4 | 19.6MB | 6.95G | 75.1 |
| +P+R | 512 | 0.4 | 18.9MB | 6.61G | 75.9 |
| +P | 512 | 0.6 | 17.1MB | 6.54G | 74.4 |
| +P+R | 512 | 0.6 | 16.3MB | 6.47G | 75.4 |
| +Direct | 300 | - | 26.1MB | 2.46G | 66.8 |
| +Direct | 512 | - | 26.1MB | 6.82G | 72.1 |
| +Mobilenet | 300 | - | 30.6MB | 2.84G | 68.1 |
| +Mobilenet | 512 | - | 30.6MB | 7.34G | 72.3 |

Lightweight object detectors with Resnet-C as backbone are trained on Pascal VOC0712 train-set with 240 epochs, while the learning rate is initialized as 0.004 and is multiplied by 0.1 at the160-th and 200-th epoch. As reported in Table 3, the created detectors (+P) with proper fixed prune-rate (0.4 or 0.6) perform well on VOC task. Moreover, adding ResBlocks (+R) to the detection branches can improve the detection accuracy. In addition, the transfer learning with knowledge distillation (+KD) helps to further improve the performance of detection models.

The SSD-Resnet50-v1d is also directly pruned and fine-tuned on VOC0712 train-set for comparison (+Direct), where the backbone is pruned using network slimming, and branch layers are not pruned. As listed in Table 3, since the feature representation capability of backbone is impaired and cannot be recovered very well by fine-tuning on small detection task, the performance of the model with '+Direct' is worse than the one with '+P'.

For further comparison, the SSD with Mobilenet-V1 as backbone is trained on VOC0712 train-set (+Mobilenet). As listed in Table 3, lightweight detectors created with '+P+R' or '+P+R+KD' outperform the SSD-Mobilenet in model size, complexity and performance, revealing favorable application value on edge devices.

## 4. CONCLUSIONS

In short, an effective model compression pipeline is designed in this paper for creating lightweight object detectors towards applications on edge devices. The proposed pipeline contains automatic channel pruning for the backbone, fixed channel deletion for the branch layers and knowledge distillation for guidance learning. Finally, the lightweight SSD performing better than SSD-Mobilenet is achieved by using the proposed compression pipeline.